\documentclass[11pt]{article}

\usepackage{fullpage}

\usepackage{microtype}
\usepackage{graphicx}
\usepackage{booktabs,bookmark} 

\usepackage{hyperref}

\usepackage{algorithm}
\usepackage{algorithmic}

\usepackage{amsfonts}       
\usepackage{nicefrac}       

\usepackage{epstopdf}
\usepackage{enumitem}
\usepackage{courier}
\usepackage{mathtools}
\usepackage{color,amsmath,amssymb,multirow}
\usepackage{multicol}

\def\vec#1{\mbox{\boldmath $#1$}}
\def\mat#1{\mbox{\bf #1}}
\usepackage{additional_def}

\newcommand{\argmin}{\mathop{\rm argmin}\limits}
\newcommand{\gradf}{{\rm grad} f}

\newcommand{\egradf}{{\rm egrad} f}

\newif\iflongversion
 \longversionfalse

\begin{document} 
\title{Riemannian joint dimensionality reduction and dictionary learning\\ on symmetric positive definite manifold}

\author{Hiroyuki Kasai\thanks{The University of Electro-Communications, Japan (e-mail: kasai@is.uec.ac.jp)}  \and Bamdev Mishra\thanks{Microsoft, India (e-mail: bamdevm@microsoft.com)}}



\maketitle

\begin{abstract}
Dictionary leaning (DL) and dimensionality reduction (DR) are powerful tools to analyze high-dimensional noisy signals. This paper presents a proposal of a novel Riemannian joint dimensionality reduction and dictionary learning (R-JDRDL) on symmetric positive definite (SPD) manifolds for classification tasks. The joint learning considers the interaction between dimensionality reduction and dictionary learning procedures by connecting them into a unified framework. We exploit a Riemannian optimization framework for solving DL and DR problems jointly.  Finally, we demonstrate that the proposed R-JDRDL outperforms existing state-of-the-arts algorithms when used for image classification tasks.
\end{abstract}

\begin{center}
\vspace*{0.3cm}
\textcolor{blue}{
{\small
Published in  European Signal Processing Conference (EUSIPCO 2018)  \cite{Kasai_EUSIPCO_2018}
}
}
\end{center}
\vspace*{0.3cm}

\section{Introduction}
\label{Sec:Introduction}
Dictionary leaning (DL) combined with sparse representation (SR) has become popular for many computer vision tasks. Many DL algorithms, e.g., K-SVD \cite{Aharon_IEEETSP_2006}, were applied originally for unsupervised learning tasks. Recently, some supervised DL algorithms have been proposed for classification tasks which exploit {\it class label} information in the training samples. They include D-KSVD \cite{Zhang_CVPR_2010} and LC-KSVD \cite{Jiang_PAMI_2013}, to name a few. However, DL for high-dimensional data is computationally expensive. To circumvent this issue, dimensionality reduction (DR) techniques are used which reduce the computational cost and highlight the low-dimensional discriminative feature of the data. 

In general, DR is applied first to the data samples, and then the dimensionality-reduced data are used for DL. The separately pre-learned DR projection matrix, however, does not fully promote the latent structure of data or preserve the best feature for DL \cite{Nguyen_ECCV_2012}. To address this issue, Feng et al. \cite{Feng_PR_2013} have proposed integration of DL and DR for improvement of the discriminative classification performance, in which a specific constraint similar to the Fisher linear discriminative analysis is imposed on the coefficient matrix. Similarly, Yang et al. \cite{Yang_MTA_2016} propose learning of the projection matrix and class-specific dictionary jointly. Li et al. \cite{Liu_BMVC_2016} report an integrated learning method of the non-negative projection matrix. Foroughi et al. \cite{Foroughi_arXiv_2016} discuss specific constraints on the coefficient matrix and on the projection matrix.

In many computer vision tasks, data of interest often reside on a {\it manifold}, which is a generalization of the Euclidean space. A particular manifold of interest is the manifold of {\it symmetric positive definite} (SPD) matrices that has been widely used in many applications. For example, region covariance matrices (RCM), which are symmetric positive definite, give good performance in texture classification and face recognition tasks \cite{Pang_CSVT_2008,Tuzel_ECCV_2006}. {The diagonal elements of a RCM represent the variances of coponent features, and the off-diagonal elements indicate the respective correlations among them. Therefore, the RCM can represent multiple features in a natural way.} It should be noted that the SPD matrices form a {\it Riemannian manifold}, which allows to understand the geometry of the space \cite{Bhatia_2007}. Cherian and Sra \cite{Cherian_2016_IEEENNLS} exploit the manifold structure to propose a Riemannian DL and sparse coding (SC) algorithm. Separately, the Riemannian DR techniques have been proposed in several works \cite{Harandi_PAMI_2017,Huang_AAAI_2017,Huang_ICML_2015,Huang_TCSVT_2017}. 

In this paper, our main contribution is to learn DL and DR jointly in the Riemannian framework. We propose R-JDRDL, an algorithm for jointly learning the projection matrix for DR and the discriminative dictionary on the SPD matrices for classification tasks. The joint learning considers the interaction between DR and DL procedures by connecting them into a unified framework. The model is formulated as an objective function over a sparse coefficient matrix and a {\it Cartesian product manifold} that consists of the Stiefel manifold and multiple SPD manifolds. Optimization on the Cartesian product manifold is cast as an optimization problem on Riemannian manifolds \cite{Absil_OptAlgMatManifold_2008}. {Optimization on the sparse coefficient matrix, on the other hand, is a convex program. }

This paper is organized as follows. 
Section II briefly introduces the SPD manifold and the Riemannian DL. 
Section III details the proposed R-JDRDL algorithm. 
Our initial results on the {MNIST image classification task} in Section IV show that R-JDRDL outperforms state-of-the-art algorithms in the domain.

\section{SPD manifold and Riemannian DL}
\label{Sec:}
This section briefly explains the geometry of SPD manifold and then introduces the Riemannian DL. Hereinafter, we denote the scalars with lower-case letters $(a, b, \ldots)$, vectors with bold lower-case letters $(\vec{a}, \vec{b}, \ldots)$, and matrices with bold-face capitals $(\mat{A}, \mat{B}, \ldots)$.  We denote a multidimensional or multi-{\it order} array as a {\it tensor}, which is denoted by $(\mathbfcal{A}, \mathbfcal{B}, \ldots)$. 

\subsection{Geometry of SPD manifold \cite{Bhatia_2007}}
\label{Subsec:}
A manifold $\mathcal{M}$ of dimensional $d$ is a topological space that locally resembles {the} Euclidean space $\mathbb{R}^d$ in a neighborhood of each point $\mat{X} \in \mathcal{M}$. All the tangent vectors at \mat{X} form a {vector} space called the tangent space of $\mathcal{M}$ at \mat{X} and denoted as $T_{\scriptsize \mat{X}}\mathcal{M}$. When endowed with a smoothly defined metric, i.e., inner product $\langle \cdot, \cdot \rangle_{\scriptsize \mat{X}}$ between vectors in the tangent space at $\mat{X} \in \mathcal{M}$, the manifold $\mathcal{M}$ is called a Riemannian manifold. The space of $d \times d$ SPD matrices, denoted as $\mathcal{S}^d_{++}$, is a Riemannian manifold, called {\it SPD manifold}, when endowed with an appropriate {\it Riemannian metric}. The tangent space at any point on $\mathcal{S}^d_{++}$ is identifiable with the set symmetric matrices $\mathcal{S}^d$. 

One particular choice of the Riemannian metric on the SPD manifold is the affine-invariant Riemannian metric (AIRM) \cite{Pennec_IJCV_2006, Bhatia_2007}. If $\mat{P}$ is an element on $\mathcal{S}^d_{++}$, the AIRM is defined as 
\vspace*{0.1cm}
\begin{equation*}
	\langle \mat{V}, \mat{W}\rangle_{\mat{P}}:=\langle \mat{P}^{-1/2}\mat{V} \mat{P}^{-1/2},  \mat{P}^{-1/2} \mat{W}\mat{P}^{-1/2} \rangle, 
	\vspace*{0.2cm}
\end{equation*}
where $\mat{V}, \mat{W} \in T_{\scriptsize \mat{P}} \mathcal{S}^d_{++}$. The choice of metric does not change with affine action by ${\rm GL}(d)$, which means that $[\mat{X}\rightarrow\mat{M}\mat{X}\mat{M}^T, \mat{X} \in \mathcal{S}^{d}_{++}, ^\forall \mat{M} \in {\rm GL}(d)]$ on $\mat{V}, \mat{W}$ and $\mat{P}$. The Riemannian metric provides a way to compute the distance between two points on the manifold. Because the SPD manifold with the AIRM metric has a unique shortest path, which is called {\it geodesic}, between every two points \cite[Section 6]{Bhatia_2007}, the geodesic distance $d:\mathcal{S}^n_{++} \times \mathcal{S}^n_{++} \rightarrow [0, \infty]$ is given as 
\vspace*{0.1cm}
\begin{equation*}
d^2(\mat{A}, \mat{B}):={\rm Log}\|\mat{A}^{-1/2}\mat{B} \mat{A}^{-1/2} \|^2_F,
\vspace*{0.1cm}
\end{equation*}
where $\mat{A}, \mat{B} \in \mathcal{S}^n_{++}$, $\| \cdot \|_F$ denotes the Frobenius norm, and ${\rm Log}$ denotes the matrix logarithm. 

\vspace*{0.2cm}
\subsection{Riemannian DL (R-DL)}
Let $\mathbfcal{X}=\{\mat{X}_1, \ldots, \mat{X}_N \} \in \mathbb{R}^{d\times d \times N}$ be the input training sample set, where $\mat{X}_n$ denotes $n$-th sample that forms a SPD matrix $\mat{X}_n \in \mathcal{S}^{d}_{++}$. The dictionary to be learned is denoted as $\mathbfcal{D} = \{ \mat{D}_1, \ldots, \mat{D}_H\} \in \prod^H \mathcal{S}^d_{++}$, where $\mat{D}_h \in \mathcal{S}^{d}_{++}$ is an atom of the dictionary. It {should be noted} that $\mathbfcal{X}$ and $\mathbfcal{D}$ are third-order tensors. We also denote a sparse coefficient vector as $\vec{a}_n \in \mathbb{R}^H_+$, which forms a coefficient matrix $\mat{A}=[\vec{a}_1,\ldots, \vec{a}_N] \in \mathbb{R}^{H \times N}_+$,  to represent a query SPD matrix $\mat{X}_n$ using the dictionary $\mathbfcal{D}$. It should also be emphasized that $\vec{a}_n$ is required to be {\it non-negative} to ensure that the resultant combination with the dictionary is positive definite. Therefore, we specifically represent a sparse {\it conic combination} of the dictionary and the coefficient vector as $\mathbfcal{D}\otimes\vec{a}_n:=\sum_{h=1}^H \vec{a}_{n,h} \mat{D}_h$ for $\vec{a}_{n,h}* \in \mathbb{R}^H_+$. 
Finally, the problem formulation is defined as
\begin{equation*}
	\min_{\scriptsize \mathbfcal{D} \in \prod^H \mathcal{S}^d_{++}, \mat{A} \in \mathbb{R}^{H \times N}_+} 
	\frac{1}{2} \sum_{n=1}^N  d^2(\mat{X}_n, \mathbfcal{D}\otimes\vec{a}_n) + R_a(\vec{a}_n) + R_D(\mathbfcal{D}),
\end{equation*}
where $R_a(\vec{a}_n)$ and $R_D(\mathbfcal{D})$ respectively represent the regularizers on the coefficient vector and the dictionary \cite{Cherian_2016_IEEENNLS}. To optimize this non-convex problem, an alternative minimization algorithm is used for the DL and the SC sub-problems. 

\section{R-JDRDL on SPD manifolds}
\label{Sec:Proposal}

\subsection{Problem formulation of R-JDRDL}
\label{Sec:Problemformulation}
Let $\mathbfcal{X}$ be the set of $N$ SPD matrices of size $m \times m$ accompanied with $K$ class labels, i.e., $\mathbfcal{X}=\{\mathbfcal{X}_1, \ldots, \mathbfcal{X}_k, \ldots, \mathbfcal{X}_K \} \in \mathbb{R}^{m\times m \times N}$, where $\mathbfcal{X}_k$ denotes the $k$-th class training samples. $\mathbfcal{X}_k$ is further composed of individual samples as $\mathbfcal{X}_k=\{\mat{X}_{k,1}, \ldots, \mat{X}_{k,n}, \ldots, \mat{X}_{k,N_k} \}$, where $\mat{X}_{k,n} \in \mathcal{S}^m_{++}$ and $N_k$ is the number of samples of the $k$-th class in the training set, i.e., $\sum_{k=1}^K N_k=N$. {Both} $\mathbfcal{X}$ and $\mathbfcal{X}_k$ are third-order tensors. The dictionary is denoted as $\mathbfcal{D} = \{ \mathbfcal{D}_1, \ldots, \mathbfcal{D}_k, \ldots, \mathbfcal{D}_K\}$, where $\mathbfcal{D}_k$ is the class-specific sub-dictionary associated with the $k$-th class. $\mathbfcal{D}_k$ is also composed as $\mathbfcal{D}_k = \{ \mat{D}_{k,1}, \ldots, \mat{D}_{k,h}, \ldots,\mat{D}_{k,{H_k}}\}$, where $H_k$ is the number of atoms of the $k$-th class sub-dictionary, and $\sum_{k=1}^K H_k=H$. 

As described earlier, the proposed R-JDRDL algorithm learns not only the dictionary $\mathbfcal{D}$, but also the projection matrix $\mat{U} \in \mathbb{R}^{m \times d} (d<m)$, which projects $m$-dimensional data {onto} $d$-dimensional data space. More specifically, $\mat{X}_{k,n} \in \mathcal{S}_{++}^m$ is mapped into $\mat{U}^T\mat{X}_{k,n} \mat{U} \in \mathcal{S}_{++}^d$. Here, we need only {\it full-rankness} of \mat{U} to guarantee that $\mat{U}^T\mat{X}_{k,n} \mat{U}$ is a SPD matrix. Equivalently, we could enforce a {\it unitary constraint} on $\mat{U}$, i.e., {$\mat{U}^T\mat{U}=\mat{I}$}.  {The space of unitary matrices is called} the {\it Stiefel manifold} {{\rm St$(d, m)\coloneqq\{ \mat{U} \in \mathbb{R}^{m \times d}: \mat{U}^T \mat{U} = \mat{I} \}$}}.

Considering that model parameters are $(\mat{U}, \mathbfcal{D}) \in \mathcal{N}$ and $\mat{A} \in \mathbb{R}^{H \times N}_+$, where $\mathcal{N}$ denotes the space of the product manifold 
$\{{\rm St}{(d,m)} \times \prod^H \mathcal{S}^d_{++}\}$, {our proposed formulation is}
\begin{eqnarray}
	\label{Eq:ProblemFormulation}
	 \{ \hat{\mat{U}}, \hat{\mathbfcal{D}}, \hat{\mat{A}} \}  &=& 
	  \argmin_{(\scriptsize \mat{U}, \mathbfcal{D}) \in \mathcal{N}, \mat{A} \in \mathbb{R}^{H\times N}_+} 
	J_d(\mat{U}, \mathbfcal{D}, \mat{A}) 
	+ \lambda_a J_a(\mat{A}) + \lambda_u J_u(\mat{U}) \nonumber\\ 
	&&\hspace*{2.9cm}+ \lambda_1 R_s(\mat{A}) + \lambda_2 R_r(\mat{A})+ \lambda_d R_d(\mathbfcal{D}),\qquad
\end{eqnarray}	
where $J_d(\mat{U}, \mathbfcal{D}, \mat{A})$ is the discriminative reconstruction error and where $J_a(\mat{A})$ and $J_u(\mat{U})$ represent the graph-based constraints on the coefficient and the projection matrices, respectively. $R_s(\mat{A})=\vec{1}_H^T  |\mat{A} | \vec{1}_N\ (:=\sum_{k=1}^K \sum_{n=1}^{N_k} \| \vec{a}_{k,n} \|_1)$, which imposes sparsity on $\mat{A}$. $R_r(\mat{A})=\| \mat{A} \|^2_F$. $\lambda$s are {non-negative regularization parameters}. $J_d$, $J_u$, and $J_a$ are described below.

{\bf Discriminative reconstruction error term {$J_d$}:}
The dictionary $\mathbfcal{D}$ is expected to approximate the dimensionality-reduced samples from all classes, of which error is represented as $d^2(\mat{U}^T\mat{X}_{k,n} \mat{U}, \mathbfcal{D}\otimes\vec{a}_{k,n})$, {where $d$ is the Riemannian geodesic distance on the SPD manifold}. In addition, to impose a more discriminative power on $\mathbfcal{D}$, the $k$-th sub-dictionary $\mathbfcal{D}_k$ is expected to approximate the dimensionality-reduced training samples associated with the $k$-th class. Here, let $\vec{a}^k_{k,n}$ be the sub-vector that corresponds to the $k$-th sub-dictionary as $\vec{a}_{k,n} = [\vec{a}^1_{k,n}; \ldots; \vec{a}^k_{k,n};\ldots; \vec{a}^K_{k,n}]$, where $\vec{a}^k_{k,n} \in \mathbb{R}^{H_k}$. The error is equivalent to $d^2(\mat{U}^T\mat{X}_{k,n} \mat{U}, \mathbfcal{D}_k\otimes\vec{a}^k_{k,n})$. It should be small. The sub-vector $\vec{a}^j_{k,n} (j \neq k)$ corresponding to other classes should be nearly zero, such that $\| \mathbfcal{D}_j\otimes\vec{a}^j_{k,n} \|_F^2$ is small. Consequently, we obtain the cost function for $J_d$ as
\begin{eqnarray}
	J_d(\mat{U}, \mathbfcal{D}, \mat{A}) 
	& \coloneqq&  \frac{1}{2}\sum_{k=1}^K  \sum_{n=1}^{N_k} (d^2(\mat{U}^T\mat{X}_{k,n} \mat{U}, \mathbfcal{D}\otimes\vec{a}_{k,n})
	 + d^2(\mat{U}^T\mat{X}_{k,n} \mat{U}, \mathbfcal{D}_k\otimes\vec{a}^k_{k,n}) )\nonumber \\
	&&+ \lambda_d \sum_{j=1, j \neq k}^K \sum_{n=1}^{N_k} \| \mathbfcal{D}_j\otimes\vec{a}^j_{k,n} \|_2^2,
\end{eqnarray}	
$\lambda_d > 0$ is the regularization parameter. 

{\bf Graph-based coefficient term {$J_a$}:}
We enforce \mat{A} to be more discriminative, and therefore, we seek to constrain the intra-class coefficients to be mutually similar and the inter-class ones to be highly dissimilar. To this end, we first construct an geometry-aware intrinsic graph of intra-class and a penalty graph for inter-class discrimination for two points $\mat{X}_p,\mat{X}_q \in \mathcal{S}_{++}^m$ as
\begin{eqnarray*}
	\label{Eq:Def_Graph_bin_G_w}
	{\mat G}_{bin}^w(p,q) & = & \left\{
	\begin{array}{ll}
		1 &\ {\rm if}\ \mat{X}_p \in N_w(\mat{X}_q) \ {\rm or\ } \mat{X}_p \in N_w(\mat{X}_q) \\
		0 &\ {\rm otherwise},
	\end{array}	
	\right.\\
	\label{Eq:Def_Graph_G_b}
	{\mat G}_{bin}^b(p,q) & = & \left\{
	\begin{array}{ll}
		1 &\ {\rm if}\ \mat{X}_p \in N_b(\mat{X}_q) \ {\rm or\ } \mat{X}_p \in N_b(\mat{X}_q) \\
		0 &\ {\rm otherwise},
	\end{array}	
	\right.	
\end{eqnarray*}	
where $N_w(\mat{X})$ is the set of $v_w$ nearest intra-class neighbors of \mat{X} in terms of geodesic distance. Similarly, $N_b(\mat{X})$ is the set of $v_b$ nearest inter-class neighbors of \mat{X}. Considering the distance of pairs of coding coefficient vectors $\vec{a}_{p}$ and $\vec{a}_{q}$ as an indicator of discrimination capability,  the final graph-based coefficient term $J_a(\mat{A})$ is defined as
\begin{eqnarray*}
	J_a(\mat{A}) &  \coloneqq & \sum_{p=1}^N \sum_{q=1}^N\frac{1}{2} \| \vec{a}_{p} - \vec{a}_{q}\|_2^2\ {\mat G}_{bin}(p,q),
\end{eqnarray*}
where ${\mat G}_{bin}(p,q) = {\mat G}_{bin}^w(p,q) - {\mat G}_{bin}^b(p,q)$ \cite{Harandi_PAMI_2017}.
This term enforces minimization of the difference of the two coding coefficients if they are the same class, although the difference of the code is maximized if they are from different classes.

{\bf Graph-based projection term {$J_u$}:}
We also learn a projection matrix $\mat{U} \in {\rm St}{(d,m)}$ that can preserve class information and which can map the training samples to a {low-dimensional} discriminative space. {Consequently, $J_u(\mat{U})$ is defined as
\begin{eqnarray*}
	J_u(\mat{U}) &  \coloneqq &  \sum_{p=1}^N \sum_{q=1}^N\frac{1}{2}  d^2(\mat{U}^T\mat{X}_{p} \mat{U}, \mat{U}^T\mat{X}_{q} \mat{U})\ {\mat G}_{rd}(p,q),
\end{eqnarray*}
where the affinity matrix ${\mat G}_{rd}$ allows to assign different weights to the Riemannian distance between different points, e.g., the distance $d(\mat{X}_p, \mat{X}_q)$ is assigned the weight ${\mat G}_{rd}(p,q)$.} 

\subsection{Optimization of R-JDRDL}
\label{Subec:Optimization}
{The objective function of (\ref{Eq:ProblemFormulation}) is divided into two sub-problems, which are solved in alternating fashion. We discuss both the sub-problems below.}

{\bf DL sub-problem on the product manifold:}
We consider the DL sub-problem of (\ref{Eq:ProblemFormulation}) by optimizing the projection matrix $\mat{U}$ and the tensor-formed dictionary $\mathbfcal{D}$, keeping \mat{A} fixed to $\hat{\mat{A}}=(\hat{\vec{a}}_{k,n})$. Consequently, the problem is can be re-formulated as
\begin{eqnarray}
	\label{Eq:ProblemFormulation_D_U}
	\min_{\scriptsize (\mat{U}, \mathbfcal{D}) \in \mathcal{N} } f(\mat{U}, \mathbfcal{D}) & \coloneqq & 
	J_d(\mat{U}, \mathbfcal{D}, \hat{\mat{A}})  + \lambda_u J_u(\mat{U}) + \lambda_d R_d(\mathbfcal{D})\nonumber \\
	& =&\frac{1}{2}\sum_{k=1}^K \sum_{n=1}^{N_k} ( d^2(\mat{U}^T\mat{X}_{k,n} \mat{U},\mathbfcal{D}\otimes\hat{\vec{a}}_{k,n})  
	  + d^2(\mat{U}^T\mat{X}_{k,n} \mat{U},\mathbfcal{D}_k\!\otimes\!\hat{\vec{a}}^k_{k,n}) )\nonumber \\
	&& + \lambda_{da}\sum_{j=1, j \neq k}^K \sum_{n=1}^{N_k} \| \mathbfcal{D}_j\!\otimes\!\hat{\vec{a}}^j_{k,n} \|_2^2 
	+\lambda_u \sum_{p=1}^N \sum_{q=1}^N\frac{1}{2}d^2(\mat{U}^T\mat{X}_{p} \mat{U}, \mat{U}^T\mat{X}_{q} \mat{U}) {\mat G}_{dr}(p,q) \nonumber\\
	&&+ \lambda_d R_d(\mathbfcal{D}).\nonumber	
\end{eqnarray}

We exploit the Riemannian optimization framework on the Cartesian product manifold $\mathcal{N}$ {(consisting of the Stiefel manifold and multiple SPD manifolds)}. In particular, we use the Riemannian conjugate gradient (RCG) method for solving the {DL sub-problem}. Theoretical convergence of the Riemannian algorithms is to a stationary point. The convergence analysis follows from \cite{Sato15a, Ring_SIAMJOptim_2012_s}. To this end, we require {the expression for the} Riemannian gradient. According to \cite{Cherian_2016_IEEENNLS}, the Riemannian gradient is obtained as $\gradf(\mat{U},\mathbfcal{D})=\mat{D}_{k,h} \egradf(\mat{U}, \mathbfcal{D}) \mat{D}_{k,h}$ with respect to $\mat{D}_{k,h}$ from the definition of AIRM where $\egradf(\mat{U},\mathbfcal{D})$ is the Euclidean gradient of $f(\mat{U},\mathbfcal{D})$ with respect to $\mat{D}_{k,h}$. 

{\bf SC sub-problem: }
We consider the SC sub-problem of (\ref{Eq:ProblemFormulation}) for solving $\mat A$, keeping \mat{U} and $\mathbfcal{D}$ fixed to $\hat{\mat{U}}$ and $\hat{\mathbfcal{D}}$, respectively. {The problem, therefore, can be re-formulated} as
\begin{eqnarray}
	\label{Eq:ProblemFormulation_A_1}
	\min_{\scriptsize \mat{A} \in \mathbb{R}_+^{H \times N} } \Psi(\mat{A})  &:=&
	J_d(\hat{\mat{U}}, \hat{\mathbfcal{D}}, \mat{A})  + \lambda_a J_a(\mat{A}) +  \lambda_1 R_s(\mat{A}) + \lambda_2 R_r(\mat{A}) \nonumber \\
	&=&\frac{1}{2} \sum_{k=1}^K (\sum_{n=1}^{N_k}
	d^2(\hat{\mat{U}}^T\mat{X}_{k,n} \hat{\mat{U}},\hat{\mathbfcal{D}}\otimes\vec{a}_{k,n})
	+ d^2(\hat{\mat{U}}^T\mat{X}_{k,n} \hat{\mat{U}}, \hat{\mathbfcal{D}}_k\!\otimes\!\vec{a}^k_{k,n}))   \nonumber \\
	&&  + \lambda_d  \sum_{j=1, j \neq k}^K \sum_{n=1}^{N_k} \| \hat{\mathbfcal{D}}_j\otimes\vec{a}^j_{k,n} \|_2^2 
	+ \sum_{p=1}^N \sum_{q=1}^N\frac{1}{2} \| \vec{a}_{p} - \vec{a}_{q}\|_2^2\ {\mat G}_{bin}(p,q)\nonumber\\
	&&  +  \lambda_1 R_s(\mat{A}) + \lambda_2 R_r(\mat{A}),\nonumber
\end{eqnarray}
where $\vec{a}_{k,n}$ is denoted as $\vec{a}_{p}$ for simplicity. Here, we calculate each column of \mat{A}, i.e., $\vec{a}_{k,n}$ sequentially by fixing the other coefficients. 

It should be emphasized that the above problem is a convex problem and is solved with a gradient projection algorithm. Specifically, we use the spectral projected gradient (SPG) solver \cite{Birgin_ACMTMS_2001,Cherian_2016_IEEENNLS}.

{\bf Classification scheme:}
{We apply the learned projection matrix $\mat{U}$ and the dictionary $\mathbfcal{D}$ on the query test sample $\mat{X}_{test}$ to estimate its class label}. For this purpose, the test sample is first projected into the low-dimensional space by $\mat{U}$. {Subsequently,} it is coded over $\mathbfcal{D}$ by solving the following equation:
\begin{equation*}
	\hat{\vec{a}}  =  {\rm arg\ min}_{\scriptsize \vec{a} \in \mathbb{R}_+^{n}} 
	\frac{1}{2}  d^2(\mat{U}^T\mat{X}_{test}\mat{U}\, \mathbfcal{D}\otimes\vec{a}) + \lambda_1 \| \vec{a}\|_1,
\end{equation*}
where $\hat{\vec{a}}=[\hat{\vec{a}}^1, \ldots, \hat{\vec{a}}^k, \ldots, \hat{\vec{a}}^K]^T$. $\hat{\vec{a}}_k$ is the sub-vector corresponding to the sub-directory $\mathbfcal{D}_k$. The residual for the $k$-th class is calculated as 
\begin{equation*}
	e_k=d^2(\mat{U}^T\mat{X}_{test}\mat{U}\, \mathbfcal{D}_k \otimes\hat{\vec{a}}^k) + \sigma \| \hat{\vec{a}} - \vec{m}_k\|^2_2,
\end{equation*}	 
where $\sigma$ is a weight to balance these two terms. $\vec{m}_k$ is the mean vector of the learned coding coefficient matrix of the $k$-th class, i.e., $\mat{A}_k$. {We adopt the distance between $\hat{\vec{a}}$ and the mean vector of the learned coding coefficient of the corresponding $k$-th class because it gives better classification results as shown in \cite{Yang_ICCV_2011}}. Finally, the identity of the testing sample is determined by selecting the class label with the minimum $e_k$.

\section{Numerical experiments}
\label{Sec:Numerica experiments}

In this section, we show the effectiveness of the proposed R-JDRDL algorithm against state-of-the-art classification algorithms on SPD matrices. 

The comparison methods are the following: NN-AIRM is the AIRM-based nearest neighbor (NN) classifier; NN-Stein is the Stein metric-based NN classifier. The Stein metric $d_S:\mathcal{S}^n_{++}\times\mathcal{S}^n_{++}\rightarrow [0,\infty]$ is a symmetric type of Bregman divergence and is defined as $d^2_S(\mat{A}, \mat{B}):= \ln \det ((\mat{A}+\mat{B})/2)+0.5\ln \det (\mat{AB})$, where $\mat{A}$ and $\mat{B} \in \mathcal{S}_{++}^d$ \cite{Sra_NIPS_2012}. DR-NN-AIRM is the AIRM-based NN classifier with the dimensionality-reduced training samples, which are obtained by R-DR \cite{Harandi_PAMI_2017}. DR-NN-AIRM is the same algorithm, but the distance metric is the Stein metric. R-SRC-AIRM and R-SRC-Stein are the sparse representation classifiers (SRCs) based on the AIRM and Stein metrics, respectively. R-KSRC stands for kernel-based SRC with the Stein metric. R-DL is the DL with the SRC classifier \cite{Cherian_2016_IEEENNLS}. R-DR-DL-AIRM and R-DR-DL-Stein are the DL with the SRC classifier after the R-DR algorithm. 

We implement our proposed algorithm in Matlab. The DL sub-problem on the product manifold makes use of the Matlab toolbox Manopt \cite{Boumal_Manopt_2014_s}. The Matlab codes R-DL, R-DR, and R-KSRC are downloaded from the respective authors' homepages. 

We use the MNIST dataset\footnote{\url{http://yann.lecun.com/exdb/mnist/.}}, which are handwritten digits of 0--9. It has 60,000 images for training and 10,000 images for testing. For this dataset, we generate $8 \times 8$ RCMs \cite{Pang_CSVT_2008}, which is computed at $(x,y)$ from the feature vector 
\begin{equation*}
\vec{f}_{x,y}=[x, y, I(x,y), |I_x|, |I_y|, |I_{xx}|, |I_{yy}|, \theta(x,y)],
\end{equation*}
{where $I(x,y)$ is the pixel value at $(x,y)$,} $I_x:=\frac{\partial I(x,y)}{\partial x}$, $I_{xx}:=\frac{\partial^2 I(x,y)}{\partial x^2}$, and $\theta(x,y):={\rm arctan}\left(\frac{|I_y|}{|I_x|}\right)$. Then, three RCMs, one from the entire image, one from the left half and one from the right, are concatenated diagonally, which produce RCM of $24 \times 24$ size for each image. We execute 10 runs under randomly selected $10$ test samples $(N)$ with $5$ and $10$ training samples. The dictionary size $H$ is equal to that of the training sample. Therefore, the case of $H=5$ represents an extreme situation. We set the parameters of the proposed algorithm, {based on cross-validation}, to $\lambda_1 =0.0001$,  $\lambda_1 =0.001$,  and $\lambda_a =0.0001$. $\lambda_u$ are $0.01$ and $0.001$ in $H=5$ and $H=10$, respectively. We also set $v_w=v_b=H-1$. The original and reduced dimensions are $m=24$ and $d=16$, respectively. We initialize $\mat{U}$ from the DR method \cite{Harandi_PAMI_2017} using single sample per class. 

The results of the classification accuracy are presented in Table \ref{tbl:MNIST_result}. The table presents superior performances of the proposed R-JDRDL against state-of-the-art algorithms. It should be noted that R-DR-DL (both with Stein and AIRM metrics) give poor performance, implying that the separately pre-learned DR projection matrix might not be optimal for the subsequent DL.

\begin{table}[t]
\vspace*{0.3cm}
\caption{Accuracy results}
\label{tbl:MNIST_result}
\begin{center}
\begin{tabular}{l c c}
\toprule
\multicolumn{1}{l }{Algorithm} & \multicolumn{2}{ c}{Accuracy (Average $\pm$ Standard deviation)} \\
\multicolumn{1}{l }{Dictionary size $(H)$} & 5 & 10 \\
\midrule
NN-AIRM & $0.464 \pm 0.0433$ & $0.551 \pm 0.0400$\\
NN-Stein & $0.469 \pm 0.0418$& $0.552 \pm  0.0426$  \\
DR-NN-AIRM & $0.598 \pm 0.0643$ &$0.619 \pm 0.0547 $\\
DR-NN-Stein & $0.591 \pm 0.0713$ &$0.618\pm 0.0531$\\
RSRC-AIRM  & $0.543 \pm  0.0464$&$0.610 \pm 0.0267$ \\
RSRC-Stein & $0.546 \pm 0.0460$ &$0.612 \pm 0.0290$ \\
R-KSRC  & $0.583 \pm 0.0392$&$0.646  \pm 0.0331$ \\
R-DL  & $0.506 \pm 0.0310$ & $0.598 \pm  0.0336$\\
R-DR-DL-AIRM& $0.434 \pm 0.0455$ & $ 0.445 \pm 0.0687$\\
R-DR-DL-Stein& $0.435 \pm  0.0481$ & $0.435 \pm 0.0610$\\
{R-JDRDL (Proposed)} & ${\bf0.617 \pm 0.0280}$ & ${\bf 0.673 \pm 0.0514}$\\
\bottomrule
\end{tabular}
\end{center}
\end{table}

\section{Conclusions}
\label{Sec:Conclusions}
We have presented a Riemannian joint framework, R-JDRDL, of performing dimensionality reduction along with discriminative dictionary learning on the set of SPD matrices for classification tasks. We formulate the joint learning as an objective function with the reconstruction error term and with the constraints on the projection matrix, the dictionary, and the sparse coefficient codes. Our numerical experiments demonstrate the good performance of jointly performing DL and DR. In particular, R-JDRDL outperforms existing state-of-the-arts algorithms {for the MNIST image classification task}. 

{Extending the framework to learning with other metrics on the SPD manifold (e.g., the Stein metric or the log-Euclidean metric) will be a topic of future research, as well as having a competitive numerical implementation with extensive evaluations on other real-world datasets.}


\section*{Acknowledgements}
H. Kasai was partially supported by JSPS KAKENHI Grant Numbers JP16K00031 and JP17H01732.

\bibliographystyle{unsrt}
\bibliography{manifold_computer_vision,sparse_coding_dic_learning,manifold_optimization}

\end{document}